# STOCHASTIC MODELING OF TAG INSTALLATION ERROR FOR ROBUST ON-MANIFOLD TAG-BASED VISUAL-INERTIAL LOCALIZATION

Kayhani, Navid[1,4], McCabe, Brenda[2], Schoellig, Angela P.[3]
[1] Ph.D. Candidate, Department of Civil and Mineral Engineering, University of Toronto, Canada
[2] Professor, Department of Civil and Mineral Engineering, University of Toronto, Canada
[3] Associate Professor, Institute for Aerospace Studies, University of Toronto, Canada
[4] navid.kayhani@mail.utoronto.ca

**Abstract:** Autonomous mobile robots, including unmanned aerial vehicles (UAVs), have received significant attention for their applications in construction. These platforms have great potential to automate and enhance the quality and frequency of the required data for many tasks such as construction schedule updating, inspections, and monitoring. Robust localization is a critical enabler for reliable deployments of autonomous robotic platforms. Automated robotic solutions rely mainly on the global positioning system (GPS) for outdoor localization. However, GPS signals are denied indoors, and pre-built environment maps are often used for indoor localization. This entails generating high-quality maps by teleoperating the mobile robot in the environment. Not only is this approach time-consuming and tedious, but it also is unreliable in indoor construction settings. Layout changes with construction progress, requiring frequent mapping sessions to support autonomous missions. Moreover, the effectiveness of vision-based solutions relying on visual features is highly impacted in low texture and repetitive areas on site. To address these challenges, we previously proposed a low-cost, lightweight tag-based visual-inertial localization method using AprilTags. Tags, in this method, are paper printable landmarks with known sizes and locations, representing the environment's quasi-map. Since tag placement/replacement is a manual process, it is subject to human errors. In this work, we study the impact of human error in the manual tag installation process and propose a stochastic approach to account for this uncertainty using the Lie group theory. Employing Monte Carlo simulation, we experimentally show that the proposed stochastic model incorporated in our on-manifold formulation improves the robustness and accuracy of tag-based localization against inevitable imperfections in manual tag installation on site.

## 1   INTRODUCTION

Automated monitoring and inspections have been extensively studied in the architecture, engineering, and construction (AEC) community. These methods need frequent and high-quality input data from the job site. Visual data have been one of the dominant data modalities used in these methods due to their information richness and low cost of collection (Mostafa and Hegazy 2021). Site personnel or hired professionals commonly capture the required images using hand-held cameras and mobile phones (Lin et al. 2021). Alternatively, automated fixed cameras installed on indoor sites or mounted on tower cranes capture real-time visual feed. Manual data collection is costly and error-prone, while stationary cameras can only cover a limited area and are ineffective indoors (Hamledari et al. 2017). Mobile robots, on the other hand, can be programmed to autonomously move the camera (or other sensors) around and are ideal for automated on-site data collection (McCabe et al. 2017).

Aerial and ground robots with different levels of autonomy have been deployed as automated data collection platforms in construction and built environments. Deploying autonomous ground robots for indoor environmental air quality (Jin et al. 2018), semantic modeling (Adán et al. 2020), and building retrofit performance evaluation (Mantha et al. 2018) are examples of the proposed automated robotic data collection solutions in built environments. Above all, camera-equipped compact aerial robots, such as rotary unmanned aerial vehicles (UAVs), can efficiently provide high-resolution images from various locations and fields of view (McCabe et al. 2017). They have shown promising potential for visual data collection in indoor (Hamledari et al. 2017; McCabe et al. 2017) and outdoor construction environments (Ham et al. 2016; Siebert and Teizer 2014). However, most of the custom-built prototypes proposed in academia (Asadi et al. 2018, 2020; Kim et al. 2018) and the cutting-edge platforms deployed in the industry (e.g., Spot ("Boston Dynamics" 2021)) are costly, limiting their scalability and applicability in practice.

One of the critical enablers for autonomy is robust global localization. Autonomous mobile robots may rely on the Global Positioning System (GPS) for outdoor localization (e.g., in (Freimuth and König 2018; Lin et al. 2021)). However, reliable localization is challenging in ever-changing, low-texture, and GPS-denied indoor construction environments (Kayhani et al. 2019). The majority of the proposed indoor localization methods rely on maps generated from data gathered with the same sensor modality used in localization. These maps are built by collecting sensory measurements via robot teleoperation. The collected data are then incorporated into a coherent environmental representation using simultaneous localization and mapping (SLAM) techniques, typically offline. However, since construction environments constantly evolve and change with project progress, frequent teleoperated mapping sessions may be required. Another challenge in these ever-changing environments is the potential loss of track due to dynamic, transparent, or temporary objects. Moreover, generating and maintaining large maps require computational and storage resources, which are highly limited, particularly in aerial robots. These technical challenges in localization in indoor construction environments make safe and reliable autonomous data collection missions difficult.

To address these challenges, we previously proposed a low-cost, versatile, lightweight visual-inertial localization method using fiducial markers such as AprilTags (Kayhani et al. 2022). AprilTags are square-shaped payload tags that provide robust data association correspondences (Olson 2011). Given that the location and size of the tags are known, this method can globally localize any platform, including inexpensive, off-the-shelf UAVs, with the minimum sensor suite of a camera and an inertial measurement unit (IMU) in real-time. The proposed formulation in (Kayhani et al. 2022) is based on an on-manifold extended Kalman filter (EKF), properly addressing the topological structure of rotations and poses in 3D space. The test results showed that our tag-based localization method could estimate the vehicle's 3D position with the accuracy of $2-5\ cm$. This method leverages construction-specific processes and practices, such as frequent indoor layout surveying and 4-dimensional building information models (4D-BIM) (Kayhani et al. 2020), to provide robust indoor localization. Tag-based visual-inertial can be applied to enable indoor localization in a wide range of applications where many vision-based techniques often face challenges due to perceptual aliasing and feature scarcity. However, the manual tag placement/replacement process may be subject to installation errors, which affect localization performance.

This work relaxes the assumption of perfectly-known 3D tags' position and orientation (i.e., pose) and considers the errors in the tag installation process using a stochastic approach. We take advantage of our previously proposed on-manifold formulation's capabilities to account for uncertainties in the input tag poses in the inertial frame, i.e., a fixed global reference frame. It is assumed that the underlying uncertainty in the determination of the tags' pose can be stochastically represented as a random variable with multi-variant zero-mean Gaussian distribution $\boldsymbol{\epsilon}_\tau \sim \mathcal{N}(\boldsymbol{0}, \boldsymbol{\Sigma}_\tau)$. In the remainder of this paper, we briefly provide some background regarding the Lie group theory and our proposed on-manifold tag-based EKF. Next, we update our formulation by incorporating uncertainty on the input global tag pose. Finally, we study the impact of incorporating the stochastic tag pose model in localization accuracy using Monte Carlo simulation and the data collected from laboratory and simulation experiments.

## 2    LIE GROUP THEORY AND STATE ESTIMATION

Motivated by the necessity of reliable performance of robotic platforms in practice, considerable effort has been devoted to the proper formulation of state estimation problems that can result in precise, consistent, and stable solutions (Solà et al. 2018). The state of a system is a set of those underlying quantities of the system, by which one can describe its intended characteristics at any snapshot (Barfoot 2017). The process of reconstructing these underlying quantities is referred to as state estimation. A state estimation technique assumes a sequence of noisy measurements/observations, a sequence of inputs to the system, and a priori system and measurement models (Barfoot 2017). These estimates are imperfect, therefore uncertain. The uncertainty in estimation can be caused by many factors, including random effects and imperfect sensors, models, and computations, that must be managed and acknowledged by the estimation framework. Therefore, a robust estimation framework needs to (1) properly formulate the underlying state; and (2) account for the sources of uncertainty. Our formulation is an effort to properly consider the manifold structure of the pose and the rotation groups in 3D and carefully deal with the representation and propagation of state and the associated uncertainties over time. To provide the necessary background, the rest of this section briefly reviews the key concepts and operations in matrix Lie groups that can be leveraged in localization.

Localization is the problem of estimating the pose of a vehicle with respect to a reference frame over time. If we fix the reference frame, the vehicle can be globally localized. Tracking the pose of a vehicle in 3D space involves estimating six degrees of freedom. Pose in 3D space is a member of the 3D special Euclidean group, $SE(3)$, represented in the form of valid $4 \times 4$ transformation matrices (Barfoot 2017):

[1] $SE(3) = \left\{ \mathbf{T} = \begin{bmatrix} \mathbf{C} & \mathbf{r} \\ \mathbf{0}^T & 1 \end{bmatrix} \in \mathbb{R}^{4 \times 4} \mid \mathbf{C} \in SO(3), \mathbf{r} \in \mathbb{R}^3 \right\}$

where $\mathbf{r}$ is the 3D translation vector, and $\mathbf{C}$ is the $3 \times 3$ matrix in the special orthogonal group of $SO(3)$ that represent rotations in 3D and is defined as:

[2] $SO(3) = \{ \mathbf{C} \in \mathbb{R}^{3 \times 3} \mid \mathbf{C}\mathbf{C}^T = \mathbf{1}, det(\mathbf{C}) = 1 \}$

Mathematically, $SE(3)$ and $SO(3)$ are differentiable and continuous (i.e., smooth) manifolds of matrix Lie groups. A smooth manifold is a curved surface without edges or spikes that can be locally approximated as a linear hyperplane (Solà et al. 2018). The smoothness guarantees a unique linear tangent space at each point on the manifold, and linear space allows for applying calculus (i.e., taking derivatives and integrals). Lie groups are smooth manifolds with the nice properties of groups, such as closure, identity, inversion, and associativity. For example, the 3D surface of a unit sphere is a smooth manifold and forms a Lie group. Every point on a unit sphere looks the same, and a unique plane exists at any point on its 3D surface. The matrix Lie groups of $SE(3)$ and $SO(3)$ have also their corresponding tangent spaces that are referred to as the Lie algebra of $se(3)$ and $so(3)$, respectively. A matrix Lie group and their corresponding Lie algebra are linked through exponential and logarithmic mappings. Using the exponential mapping from $se(3)$ to $SE(3)$, we have:

[3a] $\mathbf{T} = \exp(\boldsymbol{\xi}^\wedge) = \sum_{n=0}^{\infty} \frac{1}{n!} (\boldsymbol{\xi}^\wedge)^n , \qquad \boldsymbol{\xi} = \begin{bmatrix} \boldsymbol{\rho} \\ \boldsymbol{\phi} \end{bmatrix} \in \mathbb{R}^6$

where $\boldsymbol{\rho} \in \mathbb{R}^3$ and $\boldsymbol{\phi} \in \mathbb{R}^3$ are the translational components of the pose vector $\boldsymbol{\xi}$, $\boldsymbol{\xi}^\wedge$ is a $4 \times 4$ matrix in the Lie algebra of $se(3)$, and $\boldsymbol{\phi}^\wedge \in so(3)$ is the Lie algebra associated with $SO(3)$ and equivalent to the rotation vector ($\boldsymbol{\phi} \in \mathbb{R}^3$) expressed in the skew-symmetric matrix format. Another useful operator is the dot operator, $(.)^\odot$, which is defined as $\boldsymbol{\xi}^\wedge \mathbf{p} \equiv \mathbf{p}^\odot \boldsymbol{\xi}$ (Barfoot 2017) where $\mathbf{p}$ is expressed in the homogeneous coordinates:

$$[3\text{b}] \quad \boldsymbol{\xi}^\wedge = \begin{bmatrix} \boldsymbol{\rho} \\ \boldsymbol{\phi} \end{bmatrix}^\wedge = \begin{bmatrix} \boldsymbol{\phi}^\wedge & \boldsymbol{\rho} \\ \mathbf{0}^T & 1 \end{bmatrix} \in se(3), \quad \boldsymbol{\phi}^\wedge = \begin{bmatrix} \phi_x \\ \phi_y \\ \phi_z \end{bmatrix}^\wedge = \begin{bmatrix} 0 & -\phi_z & \phi_y \\ \phi_z & 0 & -\phi_x \\ -\phi_y & \phi_x & 0 \end{bmatrix} \in so(3)$$

$$[3\text{c}] \quad \mathbf{p}^\odot = \begin{bmatrix} sx \\ sy \\ sz \\ s \end{bmatrix}^\odot = \begin{bmatrix} \boldsymbol{\varepsilon} \\ \eta \end{bmatrix}^\odot = \begin{bmatrix} \eta \mathbf{1} & -\boldsymbol{\varepsilon}^\wedge \\ \mathbf{0}^T & \mathbf{0}^T \end{bmatrix} \in \mathbb{R}^{4 \times 6}$$

By defining the inverse of the skew-symmetric operator as $(.)^\vee$, the logarithmic mapping can be used to go in the other direction (not uniquely):

$$[4] \quad \boldsymbol{\xi} = \ln(\mathbf{T})^\vee$$

An uncertain rigid body transformation ($\mathbf{T}$) can be expressed as the combination of a noise-free nominal (i.e., mean) component ($\bar{\mathbf{T}}$) and a small, zero mean, noisy, perturbation component ($\exp(\boldsymbol{\epsilon}^\wedge)$) (Barfoot 2017). Given that the perturbation is zero-mean Gaussian, we have:

$$[5] \quad \mathbf{T} = \exp(\boldsymbol{\epsilon}^\wedge) \bar{\mathbf{T}}, \quad \boldsymbol{\epsilon} \in \mathbb{R}^6 \sim \mathcal{N}(\mathbf{0}, \boldsymbol{\Sigma})$$

This on-manifold formulation using Lie group theory allows convenient transformations of distributions through other group elements using a beneficial $6 \times 6$ linear transform called the adjoint matrix of an element of $SE(3)$ (Barfoot 2017):

$$[6] \quad \text{Ad}(\mathbf{T}) = \text{Ad}\left(\begin{bmatrix} \mathbf{C} & \mathbf{r} \\ \mathbf{0}^T & 1 \end{bmatrix}\right) = \begin{bmatrix} \mathbf{C} & \mathbf{r}^\wedge \mathbf{C} \\ \mathbf{0}^T & \mathbf{C} \end{bmatrix} \in \mathbb{R}^{6 \times 6}$$

The following section summarizes our on-manifold tag-based visual-inertial localization formulation without considering the tag installation error (Kayhani et al. 2022). Next, the errors involved in the manual tag installation process are discussed. Finally, we incorporate these errors in our formulation using the capabilities of matrix Lie groups and the theoretical concepts reviewed in this section.

## 3   TAG-BASED VISUAL-INERTIAL LOCALIZATION USING ON-MANIFOLD EKF

The state ($\mathbf{x}$) to be estimated in the localization (pose tracking) problem can be defined as:

$$[7] \quad \mathbf{x} = \left\{ \{\mathbf{r}_i^{v_0 i}, \mathbf{C}_{v_0 i}\}, \{\mathbf{r}_i^{v_1 i}, \mathbf{C}_{v_1 i}\}, \dots, \{\mathbf{r}_i^{v_K i}, \mathbf{C}_{v_K i}\} \right\} = \{\mathbf{T}_0, \mathbf{T}_1, \dots, \mathbf{T}_K\}, \quad \mathbf{T}_k \in SE(3)$$

The tag-based visual-inertial localization uses an on-manifold EKF to estimate the 3D global pose of vehicles with a camera and an IMU, including low-cost, compact UAVs. It assumes that the camera lens parameters and tag poses are known a priori. Furthermore, translational and rotational velocities ($\boldsymbol{\varpi}$) and the vehicle's initial state $\check{\mathbf{T}}_0$ are the system inputs $\mathbf{v}$. Following the perturbation scheme in Eq. [5] and given an additive perturbation, the motion model to predict the state and propagate uncertainty in time can be written as (Barfoot 2017):

$$[8] \quad \text{Nominal: } \bar{\mathbf{T}}_k = \Xi_k \bar{\mathbf{T}}_{k-1}, \quad \Xi_k = \bar{\mathbf{T}}_{v_k v_{k-1}} \in SE(3)$$

$$[9] \quad \text{Perturbation: } \delta \boldsymbol{\xi}_k = \underbrace{\text{Ad}(\Xi_k)}_{\mathbf{F}_{k-1}} \delta \boldsymbol{\xi}_{k-1} + \mathbf{w}_k, \quad \mathbf{w}_k \sim \mathcal{N}(\mathbf{0}, \mathbf{Q}_k)$$

Measurements from tag detection are then incorporated to update (correct) the predictions. From a single image, it is possible to estimate a relative pose of the detected tag with respect to the camera frame. However, instead of using the relative camera-tag pose as measurements, our formulation considers the four corresponding pixel coordinates as pixel-level measurements in a tightly coupled data fusion approach. We already showed the advantage of using pixel-level tag measurements in (Kayhani et al. 2022).

The measurement model $g(.)$ can be viewed as a combination of two non-linear functions of the state $\mathbf{x}$, $z(.)$ and $s(.)$. The 3D coordinates of the $n$-th corner point of tag $j$ expressed in camera frame $z_k^{\tau_j,n}(\mathbf{x}) = \mathbf{p}_{c_k}^{p_{\tau_j,n} c_k} = [X \quad Y \quad Z]^T$ can be written as:

[10] $z_k^{\tau_j,n}(\mathbf{x}) = \mathbf{p}_{c_k}^{p_{\tau_j,n} c_k} = \mathbf{D}^T \mathbf{T}_{cv} \mathbf{T}_{v_k i} \mathbf{p}_{\tau_j,n}$ ,

where $\mathbf{D}^T = [\mathbf{1}_3 | \mathbf{0}_{3\times 1}]$, $\mathbf{T}_{cv}$ is the vehicle to camera transform determined by calibration, $\mathbf{T}_{v_k i}$ is the state to be estimated ($\mathbf{x}$), and $\mathbf{p}_{\tau_j,n}$ is the $n$-th corner point of tag $j$ expressed in inertial coordinate frame in the homogeneous format. Given the pose of tag $j$ in the inertial frame $\mathbf{T}_{\tau_j i}$ is known as a priori, we have:

[11] $\mathbf{p}_{\tau_j,n} = \mathbf{T}_{\tau_j i}^{-1} \mathbf{P}_{\tau_j,n} = \mathbf{T}_{i\,\tau_j} \mathbf{P}_{\tau_j,n}$

where $\mathbf{P}_{\tau_j,n}$ is the homogeneous coordinates of its $n$-the corner in the tag frame $\vec{\mathcal{F}}_{\tau_j}$. In (Kayhani et al. 2022), we had the ideal assumption that $\mathbf{T}_{i\,\tau_j}$ is deterministic and subject to no uncertainty. In this work, however, we relax this assumption and are interested in investigating the impact of introducing uncertainty in $\mathbf{T}_{i\,\tau_j}$ on estimation accuracy.

The second non-linearity $s(.)$ arises from the sensor model, which is an ideal pinhole camera model. Together, the non-linear measurement model can be written as:

[12a] $\mathbf{y}_{k,j}^n = g_{k,j}^n(\mathbf{x}) = s\left(z_k^{\tau_j,n}(\mathbf{x})\right) = \begin{bmatrix} u \\ v \end{bmatrix} = \mathbf{D}_\mathbf{p} \mathbf{K} \frac{1}{Z} \begin{bmatrix} X \\ Y \\ Z \end{bmatrix} + \delta \mathbf{n}_{k,j}^n, \quad \delta \mathbf{n}_{k,j}^n \sim \mathcal{N}(\mathbf{0}, \mathbf{R}_{k,j})$

where $\mathbf{y}_{k,j}^n = [u \quad v]^T$ is the pixel coordinates of the $n$-th corner of tag $j$, observed at time $k$, projected onto the frontal image plane, $\mathbf{K}$ is the $3 \times 3$ camera intrinsic matrix, $\mathbf{D}_\mathbf{p} = [\mathbf{1}_2 | \mathbf{0}_{2\times 1}]$, and $\delta \mathbf{n}_{k,j}^n$ is an additive zero-mean Gaussian measurement noise at pixel level.

The EKF algorithm involves a prediction and a correction step in a recursive manner. The prediction step projects the current estimate of the state and the covariance (uncertainties) given the previous estimate. The correction step incorporates the measurements in the prior estimates from the prediction step and updates the associated uncertainties. To obtain the recursive steps in EKF, it is necessary to linearize the motion and observation models about the state's mean ($\bar{\mathbf{x}}$), as the operating point. In this formulation, the motion model is already linear in $\delta \boldsymbol{\xi}$. Hence, we only need to linearize the measurement model in Eq. [12a]. The generic form of the linearized motion model is as follows:

[13] $\mathbf{y}_{k,j}^n \approx g_{k,j}^n(\bar{\mathbf{x}}) + \mathbf{G}_k^{\tau_j,n} \delta \boldsymbol{\xi}_k + \delta \mathbf{n}_{k,j}^n, \quad \mathbf{G}_k^{\tau_j,n} = \mathbf{S}_k^{\tau_j,n} \mathbf{Z}_k^{\tau_j,n}$

where $\mathbf{S}_k^{\tau_j,n}$ and $\mathbf{Z}_k^{\tau_j,n}$ are the Jacobians of non-linearities $s(.)$ and $z(.)$. The sensor model's Jacobian is independent of the assumption of uncertainty in $\mathbf{T}_{i\,\tau_j}$. So, we can write:

[14a] $\mathbf{S}_k^{\tau_j,n} = \left.\frac{\partial s}{\partial z_k^{\tau_j,n}}\right|_{z_k^{\tau_j,n}(\bar{x})} = \mathbf{D}_\mathrm{p} \mathbf{K} \mathbf{S}, \qquad \mathbf{S} = \begin{bmatrix} \frac{1}{Z} & 0 & -\frac{X}{Z^2} \\ 0 & \frac{1}{Z} & -\frac{Y}{Z^2} \\ 0 & 0 & 0 \end{bmatrix}$

Preserving the assumption of deterministic $\mathbf{T}_{\tau_j i}$, we already showed (Kayhani et al. 2022):

[14b] $Z_k^{\tau_j,n} = D^T T_{cv} (\bar{T}_{v_k i} T_{i\,\tau_j} P_{\tau_j,n})^{\odot}$

In the following sections, we derive the Jacobian of the first non-linearity, that is $Z_k^{\tau_j,n}$, for uncertain world to tag transforms ($T_{\tau_j i}$).

## 4 TAG INSTALLATION ERROR: CAUSES AND MODELING

We already suggested (Kayhani et al. 2020) two main strategies for the global tag pose determination. One strategy is distributing tags in the indoor workspace and surveying their pose. The other strategy is first finding the location and orientation of the tags based on a systematic placement plan that considers factors such as localizability, safety, and cost-efficiency, and then placing the tags at the specified locations. The other factor worth considering is that the tags might need to be replaced on a construction site to guarantee their functionality over time. For example, paper-printed tags might be damaged, and those surface sprayed might need redoing. All this manual work is subject to human error that can be represented as uncertainty in $T_{i\,\tau_j}$. To model the uncertainty, we assume that $T_{i\,\tau_j}$ is perturbed by a zero-mean Gaussian distribution noise, $\epsilon_\tau \sim \mathcal{N}(0, \Sigma_\tau)$. For example, if the error for placing tags on a wall parallel to XZ plane follows a zero-mean error Gaussian distribution with a standard deviation of 2 cm in position ($\sigma_x = \sigma_z = 0.02\,m$), assuming uncorrelated errors, we can write:

[15] $\Sigma_\tau = diag(0.0004,\ 0.0,\ 0.0004,\ 0.0,\ 0.0,\ 0.0)$

However, the identification and quantification of the covariance matrix, $\Sigma_\tau$, is out of the scope of this work and depends on the adopted strategy and many other factors.

## 5 STOCHASTIC REPRESENTATION OF TAG INSTALLATION ERROR IN ON-MANIFOLD FORMULATION

Using on-manifold formulation and leveraging the properties of matrix Lie groups allow for incorporating the uncertain pose $T_{i\,\tau_j} = \{\bar{T}_{i\,\tau_j},\ \Sigma_\tau\}$ into the equations introduced in *Section 3*. Following Eq [5] the uncertain pose $T_{i\,\tau_j}$ can be represented as:

[16] $T_{i\,\tau_j} = \delta T_{i\,\tau_j} \bar{T}_{i\,\tau_j} = \exp(\epsilon_\tau^\wedge)\,\bar{T}_{i\,\tau_j},\quad \epsilon_\tau \sim \mathcal{N}(0, \Sigma_\tau),\quad \Sigma_\tau \in \mathbb{R}^{6\times 6}$

where $\epsilon_\tau \in \mathbb{R}^6$ is a vector random variable. From Eq. [10] and Eq. [11], we know that:

[17a] $z_k^{\tau_j,n}(x) = D^T T_{cv} T_{v_k i} T_{i\,\tau_j} P_{\tau_j,n} = D^T T_{cv} \exp(\delta\xi_k^\wedge)\,\bar{T}_{v_k i} \exp(\epsilon_\tau^\wedge)\,\bar{T}_{i\,\tau_j} P_{\tau_j,n}$

[17b] $z_k^{\tau_j,n}(x) \approx \underbrace{D^T T_{cv} \bar{T}_{v_k i} \bar{T}_{i\,\tau_j} P_{\tau_j,n}}_{z_k^{\tau_j,n}(\bar{x})} + \underbrace{D^T T_{cv}(\bar{T}_{v_k i}\bar{T}_{i\,\tau_j} P_{\tau_j,n})^{\odot}}_{Z_k^{\tau_j,n}} \delta\xi_k + \underbrace{D^T T_{cv} \bar{T}_{v_k i}(\bar{T}_{i\,\tau_j} P_{\tau_j,n})^{\odot}}_{E_k} \epsilon_\tau$

[18a] $g_{k,j}^n(x) = y_{k,j}^n = s\left(z_k^{\tau_j,n}(x)\right) + \delta n_{k,j}^n = s\left(z_k^{\tau_j,n}(\bar{x}) + Z_k^{\tau_j,n}\delta\xi_k + E_k \epsilon_\tau\right) + \delta n_{k,j}^n$

[18b] $g_{k,j}^n(x) \approx \underbrace{s\left(z_k^{\tau_j,n}(\bar{x})\right)}_{g_{k,j}^n(\bar{x})} + \underbrace{S_k^{\tau_j,n} Z_k^{\tau_j,n}}_{G_k^{\tau_j,n}} \delta\xi_k + \underbrace{\overbrace{S_k^{\tau_j,n} E_k}^{E_k'} \epsilon_\tau + \delta n_{k,j}^n}_{\delta N_{k,j}^n} = g_{k,j}^n(\bar{x}) + G_k^{\tau_j,n}\delta\xi_k + \delta N_{k,j}^n$

By defining $\delta N_{k,j}^n \sim \mathcal{N}(0, \mathcal{R}_{k,j})$ and for M measurements, we have:

[19] $\mathcal{R}_{k,j} = E[\delta N_{k,j}^n \delta N_{k,j}^{n\,T}] = E_k' \Sigma_\tau E_k'^T + R_{k,j},\quad \mathcal{R}_k = diag(\mathcal{R}_{k,j})\,;\ \forall j \in [1,\ M]$

Finally, the *tag installation error-aware EKF* (TIE-EKF) can be written as (Barfoot 2017):

[20a] Prediction: $\check{\mathbf{T}}_k = \mathbf{\Xi}_k \hat{\mathbf{T}}_{k-1}, \quad \check{\mathbf{P}}_k = \mathbf{F}_{k-1}\hat{\mathbf{P}}_{k-1}\mathbf{F}_{k-1}^T + \mathbf{Q}_k$

[20b] Kalman Gain: $\mathbf{K}_k = \check{\mathbf{P}}_k \mathbf{G}_k^T (\mathbf{G}_k \check{\mathbf{P}}_k \mathbf{G}_k^T + \mathcal{R}_k)^{-1}$

[20c] Correction: $\hat{\mathbf{P}}_k = (\mathbf{1} - \mathbf{K}_k \mathbf{G}_k)\check{\mathbf{P}}_k, \quad \hat{\mathbf{T}}_k = \exp\left((\mathbf{K}_k(\mathbf{y}_k - \check{\mathbf{y}}_k))^\wedge\right)\check{\mathbf{T}}_k$

where $(\hat{\cdot})$ and $(\check{\cdot})$ denote posterior (estimated) and prior (predicted) quantities, respectively, and $\mathbf{K}_k$ is the *Kalman gain*. Also, $\mathcal{R}_k$ can be thought of as the covariance of an augmented additive zero-mean measurement noise that varies with time and the vehicle's prior (predicted) state.

## 6 EXPERIMENTS

Monte Carlo simulation (MCS) is used to study the performance of our on-manifold tag-based localization with (TIE-EKF) and without (EKF) (Kayhani et al. 2022) the stochastic modeling of tag installation errors. Three experiments were conducted in simulation and laboratory settings, as summarized in Table 1. In these experiments, a low-cost, compact UAV (i.e., Parrot Bebop2) was deployed as the aerial robotic platform, and three AprilTags with known size ($0.165$ m) and pose were used (Figure 1). The trajectories were planned such that at least one tag would remain in the camera's field of view at any point in time. The UAV's camera lens parameters were previously obtained by calibration. To incorporate the installation error and equivalently tag pose uncertainty, for each iteration in MCS, we perturbed tags' pose by sampling from the pose perturbation distribution $\boldsymbol{\epsilon}_\tau \sim \mathcal{N}(\mathbf{0}, \mathbf{\Sigma}_\tau)$ and exponential mapping in Eq. [16]. Then, we estimated the vehicle's pose using TIE-EKF and EKF methods for performance comparisons. The root mean squared error (RMSE) in UAV's 3D position estimates was selected as the metric for the comparisons.

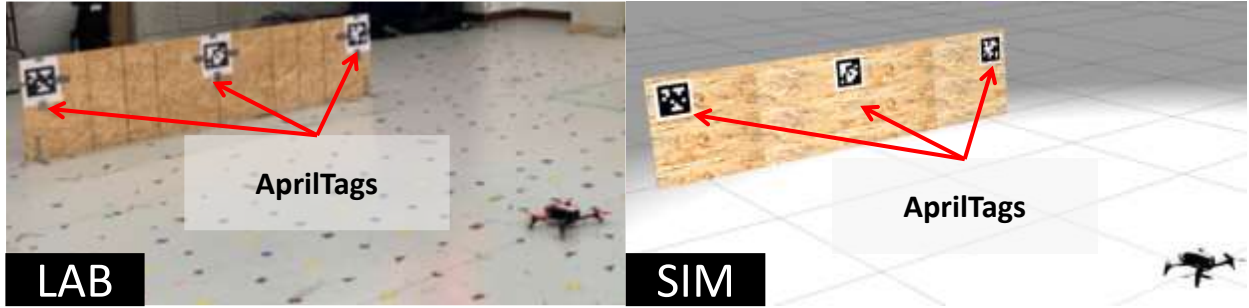

Figure 1 - Laboratory and simulation environments and the utilized tag configurations

Table 1- Summary of custom-designed experiments in laboratory and simulation environments

| Name | SIM/LAB | Description |
|---|---|---|
| SLS | SIM | A straight-line back-and-forth trajectory ($3\ m \times 4$) |
| 3DC | LAB | A 3D circular trajectory of radius one meter. |
| SLL | LAB | A straight-line back-and-forth trajectory ($3\ m \times 4$) |

In each experiment, we repeated the MCS for 200 iterations and in scenarios with different levels of uncertainty: (1) single tag with low uncertainty (*single-low*); (2) single tag with high uncertainty (*single-high*); (3) all tags with low uncertainty (*all-low*); (4) all tags with high uncertainty (*all-high*). In Scenarios 1 and 2, only the pose of the middle tag was subject to in-plane perturbation, while the other two remained untouched. In Scenarios 3 and 4, however, the poses of all the three tags were perturbed. Low and high uncertainties correspond to zero-mean Gaussian error distributions of ($\sigma_x = \sigma_z = 0.01\ m$ and $\sigma_{\theta_y} = 1°$) and ($\sigma_x = \sigma_z = 0.05\ m$ and $\sigma_{\theta_y} = 5°$), respectively.

In a separate case study, all the tag poses in experiment *3DC* were corrupted by extreme uncertainties to push the estimation methods to their limits. It is expected that highly corrupted measurements cause divergence, large errors, and biases. The reason for investigating this extreme case is to stress-test the robustness of the methods against unusual measurement uncertainties caused by errors in tag installations or pose determinations. The extreme uncertainty corresponds to a zero-mean Gaussian error distribution of ($\sigma_x = \sigma_y = \sigma_z = 0.05\ m$ and $\sigma_{\theta_x} = \sigma_{\theta_y} = \sigma_{\theta_z} = 5°$).

## 7 RESULTS

Figure 2 shows the distribution of RMSE values in position estimation using TIE-EKF and EKF methods in the three experiments introduced in Table 1 and for the degrees of uncertainty discussed above. The reported results involve 200 iterations of MCS. In all experiments, and almost for all degrees of uncertainty, the TIE-EKF methods resulted in RMSE values with a lower median, max, and spread. However, the minimum value hardly differed regardless of the method used. As seen in Figure 2, increasing uncertainty resulted in unreliable and biased estimates in all experiments. In two experiments (SLS and 3DC), a single tag with high uncertainty performed worse than when all tags were slightly corrupted, surprisingly. One possible explanation could be that their perturbation had canceled out one another. Overall, the results suggest the higher accuracy and improved robustness for TIE-EKF.

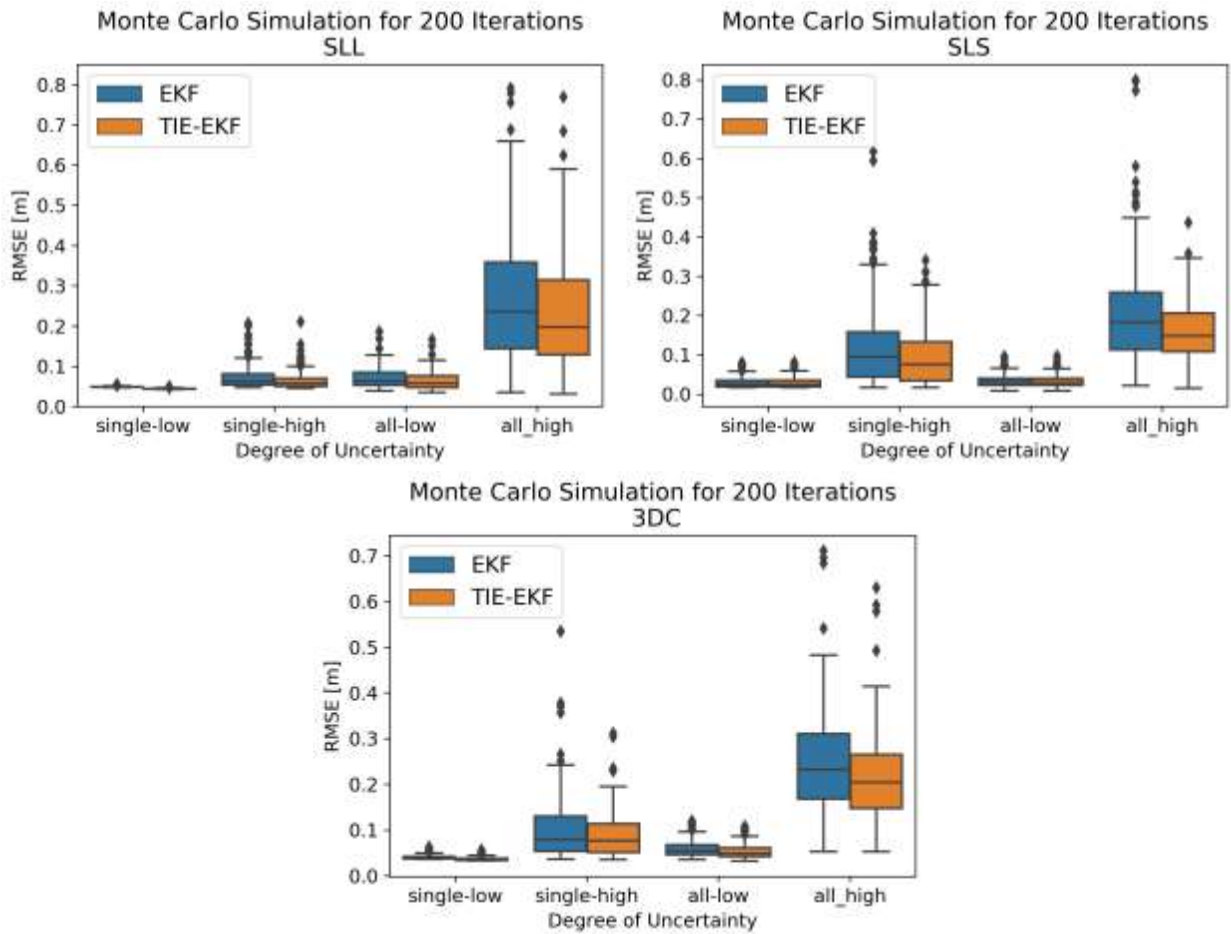

Figure 2- Comparison of TIE-EKF and EKF methods with uncertain tag poses

## 7.1 Extreme case

In *all-high* scenarios discussed above, the tag poses were subject to significant errors, already resulting in high RMSEs. However, the extreme case investigates the methods' robustness in handling even higher uncertainties in tags' pose. This test explores the estimation outcome when the tag poses are significantly off. Therefore, the high RMSE errors are expected and not a concern here, as the relative performance of the two methods is the objective of this case study.

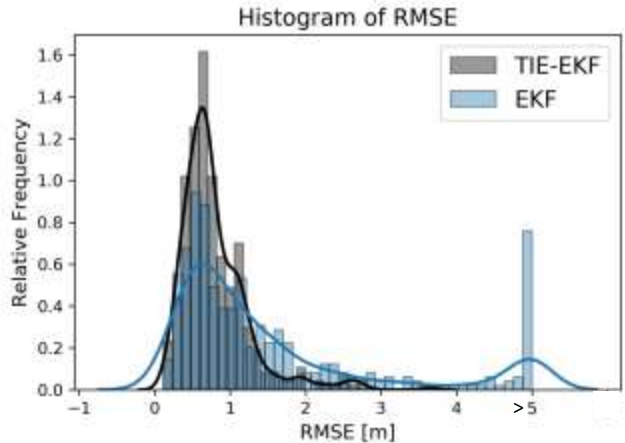

Figure 3 - RMSE histogram and density function from 400 iterations of MCS for TIE-EKF and EKF methods in an extreme tag installation error scenario in 3DC (RMSE values greater than $5\ m$ were considered as estimation divergence and consolidated in a single bin (>5))

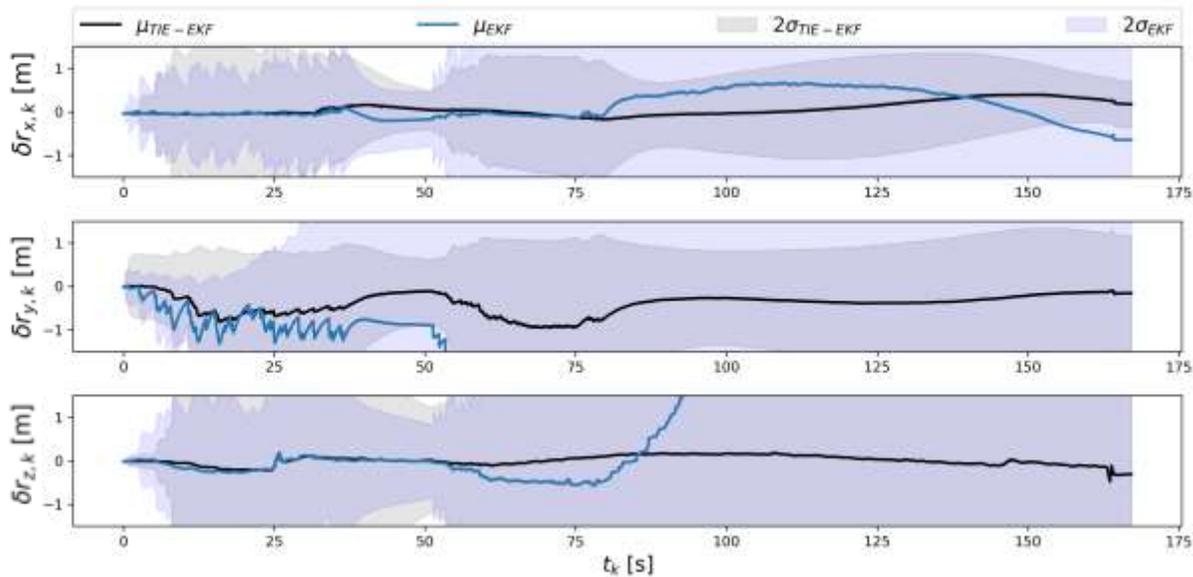

Figure 4 - Error in 3D position estimates and $2\sigma$ envelopes over time (3DC)

Figure 3 shows the histogram and density of the RMSE values for TIE-EKF and EKF in an extreme tag installation error scenario. Without considering tag installation errors (EKF), the estimates diverged and resulted in huge errors in position estimates. However, TIE-EKF had a relatively limited spread, with modes and medians closer to zero. Furthermore, Figure 4 illustrates the errors in $x$, $y$, and $z$ estimates and the $2\sigma$ envelops over time, separately. The mean error for EKF, shown in Figure 4, clearly indicates the divergence of the estimates, whereas the error values for TIE-EKF remained close to zero and did not diverge. Furthermore, the estimation variations for TIE-EKF are significantly lower than those of EKF. The results suggest that TIE-EKF is more robust even in extreme cases.

In summary, these results show that considering tag installation error and incorporating the tags' pose uncertainties in the localization formulation can improve the position estimation accuracy by 3-9%, depending on the degree of uncertainty. Furthermore, the extreme case study indicates that an error-aware estimator (TIE-EKF) is more robust against divergence in the presence of large uncertainties. In essence, TIE-EKF first acknowledges that the input tag poses might be subject to installation errors, making the tag reading measurements uncertain. Therefore, the estimator must rely less on low-quality measurements than when tag poses are more accurate. However, in the case of uncertain tag poses, since they are off from the true pose, the corresponding measurements are off, hence the corrections and consequently the final estimates. In other words, low-quality measurements will result in inaccurate estimates. This study suggests that by only acknowledging this uncertainty, the robustness and accuracy of estimates are relatively improved.

## 8   CONCLUSION

Autonomous robotic data capture solutions can enhance the efficiency of collection and the quality of required data for downstream automation tasks. Although many solutions have been proposed, they are mainly costly and face technical challenges for localization in indoor construction settings due to perceptual aliasing and feature scarcity. We previously proposed a low-cost, lightweight, versatile, tag-based visual-inertial localization method to tackle these challenges. Tags, in this method, are paper printable landmarks with known locations and sizes. Since tag placement/replacement is a manual process, it is subject to human errors. This work investigated the impact of human error in the manual tag installation process and the uncertainty in the tags' pose. Using the Lie group theory, a stochastic approach was proposed to account for this uncertainty in indoor localization. Employing Monte Carlo simulation, we experimentally showed that the proposed stochastic model incorporated in our on-manifold formulation improved the robustness and accuracy of tag-based localization against imperfections in manual tag installation on site. However, some limitations should be taken into account. First, tags were always visible in all scenarios and throughout the trajectories. Therefore, the impact of tag-blind zones should be explored in future work. Second, the distribution of tag installation errors was assumed to be zero-mean Gaussian with known statistics. More investigation might be required to determine the underlying distribution of errors and their statistics. We also recommend that future research examines the tag placement problem, considering cost and localizability.

**Acknowledgments**

Financial support from the Natural Science and Engineering Research Council (NSERC) grant number RGPIN-2017-06792 is appreciated. The first author is grateful to Prof. Tim Barfoot for his guidance and constructive advice.